\documentclass[a4paper,11pt]{article}
\usepackage[english]{babel}
\usepackage{graphicx} 
\usepackage{feynmp}
\usepackage{caption}
\usepackage{amsthm}
\usepackage{amsmath}
\usepackage{txfonts}
\usepackage{authblk}
\usepackage{abstract}
\usepackage{titlesec}
\usepackage[left=1.5in, right=1.5in]{geometry}
\usepackage{cite}
\usepackage{bbm}

\newtheorem{thm}{Theorem}[section]

\newtheorem{cor}[thm]{Corollary}
\newtheorem{prop}[thm]{Proposition}
\theoremstyle{definition}
\newtheorem{defn}[thm]{Definition}
\newtheorem{rmk}[thm]{Remark}

\newtheorem*{prf}{Proof}

\newcommand{\R}{\mathbb{R}}
\newcommand{\C}{\mathbb{C}}
\newcommand{\calX}{\mathcal{X}}
\newcommand{\calY}{\mathcal{Y}}

\newcommand{\calH}{\mathcal{H}}
\newcommand{\calN}{\mathcal{N}}
\newcommand{\rmT}{\mathrm{T}}
\newcommand{\inftyint}{\int_{-\infty}^\infty}
\newcommand{\ran}{\mathrm{Ran}}

\newcommand{\cl}{\mathrm{Cl}}
\newcommand{\diag}{\mathrm{diag}}
\newcommand{\dom}{\mathrm{Dom}}
\newcommand{\bs}{\boldsymbol}
\newcommand{\liminmean}{\mathrm{l.\,\!i.\,\!m.\ }}

\titleformat{\section}{\Large\bfseries}{\S\thesection}{1em}{}
\titleformat{\subsection}{\large\bfseries}{\S\thesubsection}{1em}{}
\titleformat{\subsubsection}{\normalsize\bfseries}{\S\thesubsubsection}{1em}{}

\title{\rule{\textwidth}{1.6pt}\vspace{-\baselineskip}\vspace{1.5pt}\\\rule{\textwidth}{0.4pt}\\{\bfseries Remarks on kernel Bayes' rule}\\\vspace{-0.4\baselineskip}\rule{\textwidth}{1pt}}
\author[1]{Hisashi Johno}
\author[2]{Kazunori Nakamoto}
\author[3]{Tatsuhiko Saigo}
\affil[1]{{\footnotesize Department of Anatomy and Cell Biology\\ Faculty of Medicine\\ University of Yamanashi\newline Shimokato 1110\\ Chuo\\ Yamanashi 409-3898\\ Japan.\quad\texttt{hisashi.johno@gmail.com}}}
\affil[2]{{\footnotesize Center for Medical Education and Sciences\\ Faculty of Medicine\\ University of Yamanashi\newline Shimokato 1110\\ Chuo\\ Yamanashi 409-3898\\ Japan.\quad\texttt{nakamoto@yamanashi.ac.jp}}}
\affil[3]{{\footnotesize Center for Medical Education and Sciences\\ Faculty of Medicine\\ University of Yamanashi\newline Shimokato 1110\\ Chuo\\ Yamanashi 409-3898\\ Japan.\quad\texttt{tsaigoh@yamanashi.ac.jp}}}
\date{}

\begin{document}
\maketitle

\begin{abstract}
\noindent
Kernel Bayes' rule has been proposed as a nonparametric kernel-based method to realize Bayesian inference in reproducing kernel Hilbert spaces.
However, we demonstrate both theoretically and experimentally that the prediction result by kernel Bayes' rule is in some cases unnatural.
We consider that this phenomenon is in part due to the fact that the assumptions in kernel Bayes' rule do not hold in general.
\end{abstract}

\section{Introduction}\label{sec:1}
\noindent
Kernel Bayes' rule has recently emerged as a novel framework for Bayesian inference \cite{song09, fukumizu13, song14}. 
It is generally agreed that, in this framework, we can estimate the kernel mean of the posterior distribution, given kernel mean expressions of the prior and likelihood distributions.
Since the distributions are mapped and nonparametrically manipulated in infinite-dimensional feature spaces called reproducing kernel hilbert spaces (RKHS), 
it is believed that kernel Bayes' rule can accurately evaluate the statistical features of high-dimensional data and enable Bayesian inference even if there were no appropriate parametric models.
To date, several applications of kernel Bayes' rule have been reported \cite{fukumizu13, kanagawa14}.
However, the basic theory and the algorithm of kernel Bayes' rule might need to be modified because of the following reasons:
\begin{enumerate}
\item\label{arg:1} 
The posterior in kernel Bayes' rule is in some cases completely unaffected by the prior.
\item\label{arg:2}
The posterior in kernel Bayes' rule considerably depends upon the choice of the parameters to regularize covariance operators.
\item\label{arg:3}
It does not hold in general that conditional expectation functions are included in the RKHS, which is an essential assumption of kernel Bayes' rule.
\end{enumerate}
This paper is organized as follows. We begin in \S\ref{sec:2} with a brief review of kernel Bayes' rule. In \S\ref{sec:3}, we theoretically address the three arguments described above. Numerical experiments are performed in \S\ref{sec:4} to confirm the theoretical results in \S\ref{sec:3}. In \S\ref{sec:5}, we summarize the theoretical and experimental results and present our conclusions. Some of the proofs for \S\ref{sec:2} and \S\ref{sec:3} are given in \S\ref{sec:6}.

\section{Kernel Bayes' rule}\label{sec:2}
\noindent
In this section, we briefly review kernel Bayes' rule following \cite{fukumizu13}.
Let $\calX$ and $\calY$ be measurable spaces, $(X, Y)$ be a random variable with an observed distribution $P$ on $\calX\times\calY$, $U$ be a random variable with the prior distribution $\Pi$ on $\calX$, and $(Z, W)$ be a random variable with the joint distribution $Q$ on $\calX\times\calY$.
Note that $Q$ is defined by the prior $\Pi$ and the family $\{P_{\calY|x}\,|\,x\in\calX\}$, where $P_{\calY|x}$ denotes the conditional distribution of $Y$ given $X=x$.
For each $y\in\calY$, let $Q_{\calX|y}$ represent the posterior distribution of $Z$ given $W = y$.
The aim of kernel Bayes' rule is to derive the kernel mean of $Q_{\calX|y}$.

\begin{defn}
Let $k_\calX$ and $k_\calY$ be measurable positive definite kernels on $\calX$ and $\calY$ such that $E[k_\calX(X, X)]<\infty$ and $E[k_\calY(Y, Y)]<\infty$, respectively, where $E[\cdot]$ denotes the expectation operator.
Let $\calH_\calX$ and $\calH_\calY$ be the RKHS defined by $k_\calX$ and $k_\calY$, respectively.
We consider two bounded linear operators $C_{YX}: \calH_\calX\rightarrow\calH_\calY$ and $C_{XX}: \calH_\calX\rightarrow\calH_\calX$ such that
\begin{equation}\label{eq:2-1}
  \langle g, C_{YX}f \rangle_{\calH_\calY} = E\left[f(X)g(Y)\right]
  \quad\mbox{and}\quad
  \langle f_1, C_{XX}f_2 \rangle_{\calH_\calX} = E\left[f_1(X)f_2(X)\right]
\end{equation}
for any $f, f_1, f_2\in\calH_\calX$ and $g\in\calH_\calY$, where $\langle \cdot, \cdot \rangle_{\calH_\calX}$ and $\langle \cdot, \cdot \rangle_{\calH_\calY}$ denote inner products on $\calH_\calX$ and $\calH_\calY$, respectively.
The integral expressions for $C_{YX}$ and $C_{XX}$ are given by 
\begin{equation*}
  (C_{YX}f)(\cdot) = \int_{\calX\times\calY} k_\calY(\cdot, y)f(x)\:dP(x, y)
  \quad \mbox{and}\quad
  (C_{XX}f)(\cdot) = \int_{\calX} k_\calX(\cdot, x)f(x)\:dP_\calX(x),
\end{equation*}
where $P_\calX$ denotes the marginal distribution of $X$.
Let $C_{XY}$ be the bounded linear operator defined by  
\begin{equation*}
  \langle f, C_{XY}g\rangle_{\calH_\calX} = E\left[f(X)g(Y) \right]
\end{equation*}
for any $f\in\calH_\calX$ and $g\in\calH_\calY$.
Then $C_{XY}$ is the adjoint of $C_{YX}$.
\end{defn}

\begin{thm}\label{thm:2-1}
\textbf{(\cite{fukumizu13}, Theorem 1)}\:
If $E[g(Y)\,|\,X=\cdot]\in\calH_\calX$ for $g\in\calH_\calY$, then $C_{XX}E[g(Y)\,|\,X=\cdot]=C_{XY}g$.
\end{thm}

\begin{defn}
Let $Q_\calY$ denote the marginal distribution of $W$.
Assuming that $E[k_\calX(U, U)]<\infty$ and $E[k_\calY(W, W)]<\infty$, we can define the kernel means of $\Pi$ and $Q_{\calY}$ by
\begin{equation*}
  m_\Pi = E[k_\calX(\cdot, U)]
  \quad\mbox{and}\quad
  m_{Q_\calY} = E[k_\calY(\cdot, W)],
\end{equation*}
respectively. Due to the reproducing properties of $\calH_\calX$ and $\calH_\calY$, the kernel means satisfy $\langle f, m_\Pi\rangle_{\calH_\calX} = E[f(U)]$ and $\langle g, m_{Q_\calY}\rangle_{\calH_\calY} = E[g(W)]$ for any $f\in\calH_\calX$ and $g\in\calH_\calY$.
\end{defn}

\begin{thm}\label{thm:2-2}
\textbf{(\cite{fukumizu13}, Theorem 2)}\:
If $C_{XX}$ is injective, $m_\Pi\in\ran({C_{XX}})$, and $E[g(Y)\,|\,X=\cdot]\in\calH_\calX$ for any $g\in\calH_\calY$, then
\begin{equation}\label{eq:2-2}
  m_{Q_\calY} = C_{YX}C_{XX}^{-1}m_\Pi,
\end{equation}
where $\ran(C_{XX})$ denotes the range of $C_{XX}$.
\end{thm}

Here we have, for any $x\in\calX$, 
\begin{equation}\label{eq:2-3}
  E\left[k_\calY(\cdot, Y)\mid X=x\right] = C_{YX}C_{XX}^{-1}k_\calX(\cdot, x)
\end{equation}
by replacing $m_\Pi$ in Equation (\ref{eq:2-2}) for $k_\calX(\cdot, x)$.
It is noted in \cite{fukumizu13} that the assumption $m_\Pi\in\ran(C_{XX})$ does not hold in general. 
In order to remove this assumption, $(C_{XX}+\epsilon I)^{-1}$ has been suggested to be used instead of $C_{XX}^{-1}$, where $\epsilon$ is a regularization constant and $I$ is the identity operator.
Thus, the approximations of Equations (\ref{eq:2-2}) and (\ref{eq:2-3}) are respectively given by
\begin{equation*}
  m_{Q_\calY}^{reg} = C_{YX}\left(C_{XX}+\epsilon I\right)^{-1}m_\Pi
  \quad\mbox{and}\quad
  E^{reg}\left[k_\calY(\cdot, Y)\mid X=x\right] = C_{YX}\left(C_{XX}+\epsilon I\right)^{-1}k_\calX(\cdot, x).
\end{equation*}
Similarly, for any $y\in\calY$, the approximation of $m_{Q_{\calX|y}}$ is provided by
\begin{equation}\label{eq:2-4}
  m_{Q_{\calX|y}}^{reg}
  = E^{reg}\left[k_\calX(\cdot, Z)\mid W=y\right]
  = C_{ZW}\left(C_{WW}+\delta I\right)^{-1}k_\calY(\cdot, y),
\end{equation}
where $\delta$ is a regularization constant and the linear operators $C_{ZW}$ and $C_{WW}$ will be defined below.

\begin{defn}
We consider the kernel means $m_Q = m_{(ZW)}$ and $m_{(WW)}$ such that
\begin{equation*}
  \langle m_{(ZW)},\, g\otimes f\rangle_{\calH_\calY\otimes\calH_\calX} = E\left[f(Z)g(W)\right]
  \quad\mbox{and}\quad
  \langle m_{(WW)},\, g_1\otimes g_2\rangle_{\calH_\calY\otimes\calH_\calY} = E\left[g_1(W)g_2(W)\right]
\end{equation*}
for any $f\in\calH_\calX$ and $g, g_1, g_2\in\calH_\calY$, where $\otimes$ denotes the tensor product.
Let $C_{(YX)X}: \calH_\calX\rightarrow\calH_\calY\otimes\calH_\calX$ and $C_{(YY)X}: \calH_\calX\rightarrow\calH_\calY\otimes\calH_\calY$ be bounded linear operators which respectively satisfy
\begin{align}\label{eq:2-5}
  \begin{split}
    \langle g\otimes f,\, C_{(YX)X}h\rangle_{\calH_\calY\otimes\calH_\calX} &= E\left[g(Y)f(X)h(X)\right], \\
    \langle g_1\otimes g_2,\, C_{(YY)X}f\rangle_{\calH_\calY\otimes\calH_\calY} &= E\left[g_1(Y)g_2(Y)f(X)\right]
  \end{split}
\end{align}
for any $f, h\in\calH_\calX$ and $g, g_1, g_2\in\calH_\calY$.
\end{defn}

From Theorem \ref{thm:2-2}, Fukumizu et al.\cite{fukumizu13} proposed that $m_{(ZW)}$ and $m_{(WW)}$ can be given by
\begin{equation*}
  m_{(ZW)} = C_{(YX)X}C_{XX}^{-1}m_\Pi\in\calH_\calY\otimes\calH_\calX
  \quad\mbox{and}\quad
  m_{(WW)} = C_{(YY)X}C_{XX}^{-1}m_\Pi\in\calH_\calY\otimes\calH_\calY.
\end{equation*}
In case $m_\Pi$ is not included in $\ran(C_{XX})$, they suggested that $m_{(ZW)}$ and $m_{(WW)}$ could be approximated by
\begin{equation*}
  m_{(ZW)}^{reg} = C_{(YX)X}\left(C_{XX}+\epsilon I\right)^{-1}m_\Pi
  \quad\mbox{and}\quad
  m_{(WW)}^{reg} = C_{(YY)X}\left(C_{XX}+\epsilon I\right)^{-1}m_\Pi.
\end{equation*}

\begin{rmk}\label{prop:2-1}
\textbf{(\cite{fukumizu13}, page 3760)}\:
$m_{(ZW)}$ and $m_{(WW)}$ can respectively be identified with $C_{ZW}$ and $C_{WW}$.
\end{rmk}

Here, we introduce the empirical method for estimating the posterior kernel mean $m_{Q_{\calX|y}}$ following \cite{fukumizu13}.

\begin{defn}
Suppose we have an independent and identically distributed (i.i.d.\ ) sample $\{(X_i, Y_i)\}_{i=1}^n$ from the observed distribution $P$ on $\calX\times\calY$ and a sample $\{U_j\}_{j=1}^l$ from the prior distribution $\Pi$ on $\calX$. The prior kernel mean $m_\Pi$ is estimated by
\begin{equation}\label{eq:2-6}
  \widehat{m}_\Pi = \sum_{j=1}^l \gamma_jk_\calX(\cdot, U_j), 
\end{equation}
where $\gamma_1,\ldots,\gamma_l$ are weights.
Let us put $\widehat{\bs{m}}_\Pi = (\widehat{m}_\Pi(X_1),\ldots, \widehat{m}_\Pi(X_n))^{\rmT}$,
$G_X = (k_\calX(X_i, X_j))_{1\le i,j\le n}$, and $G_Y = (k_\calY(Y_i, Y_j))_{1\le i,j\le n}$.
\end{defn}

\begin{prop}\label{prop:rev}
\textbf{(\cite{fukumizu13}, Proposition 3, revised)}\:
Let $I_n$ denote the identity matrix of size $n$.
The estimates of $C_{ZW}$ and $C_{WW}$ are given by
\begin{equation*}
  \widehat{C}_{ZW} = \sum_{i=1}^n\widehat{\mu}_ik_\calX(\cdot,X_i)\otimes k_\calY(\cdot,Y_i)
  \quad\mbox{and}\quad
  \widehat{C}_{WW} = \sum_{i=1}^n\widehat{\mu}_ik_\calY(\cdot,Y_i)\otimes k_\calY(\cdot,Y_i), 
\end{equation*}
respectively, where $\widehat{\bs{\mu}} = (\widehat{\mu}_1, \ldots, \widehat{\mu}_n)^\rmT = (G_X+n\epsilon I_n)^{-1}\widehat{\bs{m}}_\Pi$.
\end{prop}

The proof of this revised proposition is given in \S\ref{sec:rev_prf}. It is suggested in \cite{fukumizu13} that Equation (\ref{eq:2-4}) can be empirically estimated by
\begin{equation*}
  \widehat{m}_{Q_{\calX|y}} = \widehat{C}_{ZW}\left(\widehat{C}_{WW}^{\ 2}+\delta I_n\right)^{-1}\widehat{C}_{WW}k_\calY(\cdot, y).
\end{equation*}

\begin{thm}\label{thm:kbr-prop4}
\textbf{(\cite{fukumizu13}, Proposition 4)}\:
Given an observation $y\in\calY$, $\widehat{m}_{Q_{\calX|y}}$ can be calculated by
\begin{equation*}
  \widehat{m}_{Q_{\calX|y}} = \bs{k}_X^\rmT R_{X|Y}\bs{k}_Y(y),
  \quad
  R_{X|Y} = \Lambda G_Y\left((\Lambda G_Y)^2+\delta I\right)^{-1}\Lambda,
\end{equation*}
where $\Lambda = \diag(\,\widehat{\bs{\mu}}\,)$ is the diagonal matrix with the elements of\, $\widehat{\bs{\mu}}$, $\bs{k}_X = \left(k_\calX(\cdot,X_1),\ldots,k_\calX(\cdot,X_n)\right)^\rmT$, and $\bs{k}_Y = \left(k_\calY(\cdot,Y_1),\ldots,k_\calY(\cdot,Y_n)\right)^\rmT$.
\end{thm}

If we want to know the posterior expectation of a function $f\in\calH_\calX$ given an observation $y\in\calY$, it is estimated by
\begin{equation*}
  \langle f, \widehat{m}_{Q_{\calX|y}}\rangle_{\calH_\calX}
  = \bs{f}_X^\rmT R_{X|Y}\bs{k}_\calY(y),
\end{equation*}
where $\bs{f}_X = (f(X_1), \ldots, f(X_n))^\rmT$.

\section{Theoretical arguments}\label{sec:3}
\noindent
In this section, we theoretically support the three arguments raised in \S\ref{sec:1}.  
First, we show in \S \ref{subsec:3-1} that the posterior kernel mean 
$\widehat{m}_{Q_{\calX|y}}$ is completely unaffected by the prior distribution 
$\Pi$ under the condition that $\Lambda$ and  $G_Y$ are non-singular.  
This implies that, at least in some cases, $\Pi$ does not properly affect $\widehat{m}_{Q_{\calX|y}}$. 
Second, we mention in \S \ref{subsec:3-2} that the linear operators $C_{XX}$ and $C_{WW}$ are not always surjective, and address the problems associated with the setting of the regularization parameters $\epsilon$ and $\delta$.
Third, we demonstrate in \S \ref{subsec:3-3} that conditional expectation functions are not generally contained in the RKHS, which means that Theorems 1, 2, 5, 6, 7, and 8 in \cite{fukumizu13} do not work in some situations. 

\subsection{Relations between the posterior $\widehat{m}_{Q_{\calX|y}}$ and the prior $\Pi$}\label{subsec:3-1} 
\noindent
Let us review Theorem \ref{thm:kbr-prop4}. 
Assume that $G_Y$ and $\Lambda$ are non-singular matrices. (This assumption is not so strange, as shown in $\S\ref{sec:nonsing}$.)
The matrix $R_{X|Y} = \Lambda G_Y((\Lambda G_Y)^2+\delta I)^{-1}\Lambda$ tends to $G_Y^{-1}$ as $\delta$ tends to $0$.  
Furthermore, if we set $\delta = 0$ from the beginning, we obtain $R_{X|Y} = G_Y^{-1}$. This implies that the posterior kernel mean $\widehat{m}_{Q_{\calX|y}} = \bs{k}_X^\rmT R_{X|Y}\bs{k}_Y(y)$ never depends on the prior distribution $\Pi$ on $\calX$, which seems to be a contradiction to the nature of Bayes' rule.   
This result is numerically confirmed in \S\ref{subsec:4-2}. 

\subsection{The inverse of the operators $C_{XX}$ and $C_{WW}$}\label{subsec:3-2}
\noindent
As noted by Fukumizu et al. \cite{fukumizu13}, the linear operators $C_{XX}$ and $C_{WW}$ are not surjective in some usual cases, the proof of which is given in \S\ref{sec:nonsur}. 
Therefore, they proposed an alternative way of obtaining a solution $f\in\calH_\calX$ of the equation $C_{XX}f=m_{\Pi}$, that is, a regularized inversion $f = (C_{XX}+\epsilon I)^{-1}m_{\Pi}$ as an analog of ridge regression, where $\epsilon$ is a regularization parameter and $I$ is an identity operator.
One of the disadvantages of this method is that the solution $f = (C_{XX}+\epsilon I)^{-1}m_{\Pi}$ depends upon the choice of $\epsilon$. 
In \S \ref{subsec:4-2}, we numerically show that the prediction using kernel Bayes' rule considerably depends on the regularization parameters $\epsilon$ and $\delta$. 
Theorems 5, 6, 7, and 8 in \cite{fukumizu13} seem to support the appropriateness of 
the regularized inversion.   
However, these theorems work under the condition that conditional expectation functions are contained in the RKHS, which does not hold in several cases as proved in \S \ref{subsec:3-3}. 
Furthermore, since we need to assume sufficiently slow decay of the regularization constants $\epsilon$ and $\delta$ in these theorems, it is practically difficult to set appropriate values for $\epsilon$ and $\delta$.  
A cross-validation procedure seems to be useful for tuning the parameters and we may obtain good experimental results, however, it seems to lack theoretical background.   
  
Instead of the regularized inversion method, we can compute generalized inverse matrices of $G_X$ and $\Lambda G_Y$, given a sample $\{ (X_i, Y_i) \}_{i=1}^n$.
Below, we briefly introduce a generalization of a matrix inverse.
For more details, see \cite{horn13}. 

\begin{defn}
Let $A$ be a matrix of size $m \times n$ over the complex number space $\C$. 
We say that a matrix $A^{\times}$ of size $n \times m$ is a generalized inverse matrix of $A$ if  $AA^{\times}A=A$. 
We also say that a matrix $A^{\dag}$ of size $n \times m$ is the Moore-Penrose generalized inverse matrix of $A$ if $AA^{\dag}$ and $A^{\dag}A$ are Hermitian, $AA^{\dag}A=A$, and $A^{\dag}AA^{\dag}=A^{\dag}$. 
\end{defn} 

\begin{rmk} 
In fact, any matrix $A$ has the Moore-Penrose generalized inverse matrix $A^{\dag}$. 
Note that $A^{\dag}$ is uniquely determined by $A$. If $A$ is square and non-singular,        
then $A^{\times} = A^{\dag} = A^{-1}$. 
For a generalized inverse matrix $A^{\times}$ of size $n\times m$, $AA^{\times}v=v$ for any vector $v \in\C^m$ if $v$ is contained in the image of $A$. 
In particular, $A^{\times}v$ is a vector contained in the preimage of $v$ under $A$. 
\end{rmk}

In the calculation of $\widehat{m}_{Q_{\calX|y}} = \bs{k}_X^\rmT R_{X|Y}\bs{k}_Y(y)$, we numerically compare the case $R_{X|Y} = (\Lambda' G_Y)^{\dag}\Lambda'$ with the original case $R_{X|Y} = \Lambda G_Y((\Lambda G_Y)^2+\delta I)^{-1}\Lambda$ in \S \ref{subsec:4-2}, where $\Lambda' = \diag(G_X^\dagger\widehat{\bs{m}}_\Pi)$.

\subsection{Conditional expectation functions and RKHS}\label{subsec:3-3}
\noindent
In this subsection, we show that conditional expectation functions are in some cases not contained in the RKHS.
\begin{defn}
For $p\in[1, \infty)$, we define the spaces $L^p(\R)$, $L^p(\R, \C)$, and $L^p(\R^2, \R)$ as 
\begin{eqnarray*}
  L^p(\R) & := & \left\{\begin{array}{c|c}
    f:\R\rightarrow\R  & \displaystyle \inftyint  |f(x)|^p\:dx<\infty 
  \end{array}\right\},  \\ 
  L^p(\R, \C) & := & \left\{\begin{array}{c|c}
    f:\R\rightarrow\C  & \displaystyle \inftyint  |f(x)|^p\:dx<\infty 
  \end{array}\right\}, \\
  L^p(\R^2, \R) & := & \left\{\begin{array}{c|c}
    f:\R^2\rightarrow\R  & \displaystyle \int_{\R^2}  |f(x_1, x_2)|^p\:dx_1dx_2<\infty 
  \end{array}\right\}.
\end{eqnarray*}
We also define the $L^p$ norm for $f\in L^p(\R)$ or $f\in L^p(\R, \C)$ as
\begin{equation*}
  \|f\|_p := \left(\inftyint  |f(x)|^p\: dx \right)^{\frac{1}{p}},
\end{equation*}
and the $L^p$ norm for $f\in L^p(\R^2, \R)$ as
\begin{equation*}
  \|f\|_p := \left(\int_{\R^2}  |f(x_1, x_2)|^p\:dx_1dx_2 \right)^{\frac{1}{p}}.
\end{equation*}
\end{defn}

\begin{defn}\label{defn:FT1} 
For a function $f \in L^{1}(\R, \C) \cap L^{2}(\R, \C)$, we define its Fourier transform as
\begin{equation*}
  \hat{f}(t) := \frac{1}{\sqrt{2\pi}} \inftyint f(x) \exp(-\sqrt{-1}tx) \: dx. 
\end{equation*} 
We can uniquely extend the Fourier transform to an isometry \,
$\hat{} : L^{2}(\R, \C) \to L^{2}(\R, \C)$.  
We also define the inverse Fourier transform \, 
$\check{} : L^{2}(\R, \C) \to L^{2}(\R, \C)$ as an isometry uniquely determined by 
\begin{equation*}
  \check{f}(t) := \frac{1}{\sqrt{2\pi}} \int_{-\infty}^{\infty} f(x) \exp(\sqrt{-1}tx) \: dx 
\end{equation*} 
for $f \in L^{1}(\R, \C) \cap L^{2}(\R, \C)$. 
\end{defn} 

\begin{defn}
Let us define a Gaussian kernel $k_G$ on $\R$ by 
\begin{equation*}
  k_G(x, y) := \frac{1}{\sqrt{2\pi} \sigma} \exp \left( -\frac{(x-y)^2}{2\sigma^2} \right).
\end{equation*}
As described in \cite{fukumizu10}, the RKHS of real-valued functions and complex-valued functions corresponding to the positive definite kernel $k_G$ are given by 
\begin{align*}
  \calH_G  &:= \left\{\begin{array}{c|c}
    f\in L^2(\R) &  \displaystyle \int_{-\infty}^{\infty}\left|\hat{f}(t)\right|^2 \exp\left(\frac{\sigma^2}{2} t^2 \right)\:dt < \infty
  \end{array} \right\}, \\
  \calH_{G}\left(\R, \C  \right) &:= \left\{\begin{array}{c|c} 
     f\in L^2(\R, \C) &  \displaystyle \int_{-\infty}^{\infty}\left|\hat{f}(t)\right|^2 \exp \left(\frac{\sigma^2}{2} t^2 \right)\:dt < \infty
  \end{array} \right\},
\end{align*}
respectively, and the inner product of $f, g \in \calH_G $ or $f, g \in \calH_G \left(\R, \C  \right)$ on the RKHS is calculated by 
\begin{align}\label{eq:IP1} 
  \begin{split}
    \langle f, g \rangle = \int_{-\infty}^{\infty} \hat{f}(t) \overline{\hat{g}(t)} \exp\left(\frac{\sigma^2}{2} t^2 \right) \: dt,
  \end{split}
\end{align}
where the overline denotes the complex conjugate.
Remark that $\calH_{G}$ is a real Hilbert subspace contained in the complex Hilbert space $\calH_{G}(\R, \C)$. 
\end{defn} 

Fukumizu et al. \cite{fukumizu13} mentioned that the conditional expectation function $E[g(Y)\,|\,X=\cdot ]$ is not always included in $\calH_{\calX}$. 
Indeed, if the variables $X$ and $Y$ are independent, then $E[g(Y)\,|\, X=\cdot ]$ becomes a constant function on $\calX$, the value of which might be non-zero. 
In the case that $\calX=\R$ and $k_\calX=k_G$, the constant function with non-zero value is not contained in $\calH_\calX=\calH_G $.

Additionally, in order to prove Theorems 5 and 8 in \cite{fukumizu13}, they made the assumption that  
$E[k_{\calY}(Y, \tilde{Y})\,|\,X=x, \tilde{X}=\tilde{x}] \in \calH_{\calX}\otimes\calH_{\calX}$ and $E[k_{\calX}(Z, \tilde{Z})\,|\, W=y, \tilde{W}=\tilde{y}] \in \calH_{\calY}\otimes\calH_{\calY}$, 
where $(\tilde{X}, \tilde{Y})$ and 
$(\tilde{Z}, \tilde{W})$ are independent copies of the random variables $(X, Y)$ and $(Z, W)$ on 
$\calX \times \calY$, respectively. 
We also see that this assumption does not hold in general. 
Suppose that $X$ and $Y$ are independent and that so are $\tilde{X}$ and $\tilde{Y}$. 
Then $E[k_{\calY}(Y, \tilde{Y}) \mid X=x, \tilde{X}=\tilde{x}]$ is a constant function of $(x, \tilde{x})$, the value of which might be non-zero. 
In the case that $\calX = \R$ and $k_\calX=k_G$, the constant function having non-zero value is not contained in $\calH_{\calX}\otimes\calH_{\calX}=\calH_{G}\otimes\calH_{G}$. 
Note that $\calH_{G}\otimes\calH_{G}$ is isomorphic to the RKHS corresponding to 
the kernel $k((x_1, x_2), (\tilde{x}_1, \tilde{x}_2)) = k_G(x_1, \tilde{x}_1)k_G(x_2, \tilde{x}_2)$ on $\R^2$, that is,
\begin{equation*}
  \calH_{G}\otimes \calH_{G} = \left\{
  \begin{array}{c|c} 
   f\in L^2(\R^2, \R) &  \displaystyle \int_{\R^2} 
   \left|\hat{f}(t_1, t_2)\right|^2 
   \exp \left(\frac{\sigma^2}{2} (t_1^2+t_2^2) \right)\:dt_1dt_2 < \infty
  \end{array} \right\},  
\end{equation*}
where the Fourier transform of $f:\R^2\rightarrow \R$ is defined by  
  \begin{equation*}
    \hat{f}(t_1, t_2) := \underset{n\to\infty\;\;}{\liminmean}\left(\frac{1}
    {\sqrt{2\pi}}\right)^2 \int_{x_1^2+x_2^2 < n} f(x_1, x_2) 
    \exp\left(-\sqrt{-1}(t_1x_1+t_2x_2)\right) \: dx_1dx_2.  
  \end{equation*}  
  
Thus, the assumption that conditional expectation functions are included in the RKHS does not hold in general.
Since most of the theorems in \cite{fukumizu13} require this assumption, kernel Bayes' rule may not work in several cases.

\section{Numerical experiments}\label{sec:4}
\noindent
In this section, we perform numerical experiments to illustrate the theoretical results in \S\ref{subsec:3-1} and \S\ref{subsec:3-2}. 
We first introduce probabilistic classifiers in \S\ref{subsec:4-1} based on conventional Bayes' rule assuming Gaussian distributions (BR), original kernel Bayes' rule (KBR1), and kernel Bayes' rule using Moore-Penrose generalized inverse matrices (KBR2).
In \S\ref{subsec:4-2}, we apply the three classifiers to a binary classification problem with computer-simulated data sets.
Numerical experiments are implemented in version 2.7.6 of the Python software (Python Software Foundation, Wolfeboro Falls, NH, USA).

\subsection{Algorithms of the three classifiers, BR, KBR1, and KBR2}\label{subsec:4-1}
\noindent
Let $(X, Y)$ be a random variable with a distribution $P$ on $\calX\times\calY$, where $\calX = \{C_1,\ldots,C_g\}$ is a family of classes and $\calY = \R^d$.
Let $\Pi$ and $Q$ be the prior and the joint distributions on $\calX$ and $\calX\times\calY$, respectively. 
Suppose we have an i.i.d.\  training sample $\{(X_i, Y_i)\}_{i=1}^n$ from the distribution $P$.
The aim of this subsection is to derive algorithms of the three classifiers, BR, KBR1, and KBR2, which respectively calculate the posterior probability for each class given an observation $y\in\calY$, that is, $Q_{\calX|y}(C_1),\ldots, Q_{\calX|y}(C_g)$.

\subsubsection{The algorithm of BR}
\noindent
In BR, we estimate the posterior probability of $j$-th class $(j=1,\ldots,g)$ given a test value $y\in\calY$ by 
\begin{equation*}
  \widehat{Q}_{\calX|y}(C_j) = \frac{\widehat{P}_{\calY|C_j}(y)\:\Pi(C_j)}{\sum_{k=1}^g\widehat{P}_{\calY|C_k}(y)\:\Pi(C_k)},
\end{equation*}  
where $\widehat{P}_{\calY|C_j}(\cdot)$ is the density function of the $d$-dimensional normal distribution $\calN(\widehat{M}_j, \widehat{S}_j)$ defined by
\begin{equation*}
  \widehat{P}_{\calY|C_j}(\cdot) = \frac{1}{\sqrt{(2\pi)^d\left|\,\widehat{S}_j\right|}}
  \exp\left(-\frac{1}{2}(\,\cdot-\widehat{M}_j)^\rmT\, \widehat{S}_j^{-1}(\,\cdot-\widehat{M}_j)\right).
\end{equation*}
The mean vector $\widehat{M}_j\in\R^d$ and the covariance matrix $\widehat{S}_j\in\R^d\times\R^d$ are calculated from the training data of the class $C_j$.

\subsubsection{The algorithm of KBR1}\label{subsubsec:4-1-2}
\noindent
Let us define positive definite kernels $k_\calX$ and $k_\calY$ as
\begin{equation*}
  k_\mathcal{X}(X, X') = \begin{cases}
  \ \ 1 & (X=X') \\
  \ \ 0 & (X\ne X')\end{cases}
  \quad\mbox{and}\quad 
  k_\mathcal{Y}(Y, Y') = \frac{1}{\sqrt{2\pi}\sigma} 
  \exp \left(-\frac{\left\|Y-Y'\right\|^2}{2\sigma^2} \right)
\end{equation*}
for $X, X'\in\calX$ and $Y, Y'\in\calY$, and the corresponding RKHS as $\calH_\calX$ and $\calH_\calY$, respectively.
Here we set $\left\| Y \right\| = \sqrt{\sum_{i=1}^d y_i^2}$ for $Y = (y_1, y_2, \ldots, y_d)^{\rm T} \in \calY = \R^d$. 
Then, the prior kernel mean is given by
\begin{equation*}
  \widehat{m}_\Pi(\cdot) = \sum_{j=1}^g\Pi(C_j)k_\calX(\cdot,C_j),
\end{equation*}
where $\sum_{j=1}^g\Pi(C_j)=1$.
Let us put $G_X = (k_\calX(X_i, X_j))_{1\le i,j\le n}$, $G_Y = (k_\calY(Y_i, Y_j))_{1\le i,j\le n}$, $D = (\mathbbm{1}_{\{C_i\}}(X_j))_{1\le i\le g,1\le j\le n}\in\{0,1\}^{g\times n}$,
$\widehat{\bs{m}}_\Pi = (\widehat{m}_\Pi(X_1),\ldots, \widehat{m}_\Pi(X_n))^{\rmT}$, $\widehat{\bs{\mu}} = (\widehat{\mu}_1, \ldots, \widehat{\mu}_n)^\rmT = (G_X+n\epsilon I_n)^{-1}\widehat{\bs{m}}_\Pi$,
$\Lambda = \diag(\,\widehat{\bs{\mu}}\,)$, $\bs{k}_X(\cdot) = \left(k_\calX(\cdot,X_1),\ldots,k_\calX(\cdot,X_n)\right)^\rmT$, $\bs{k}_Y(\cdot) = \left(k_\calY(\cdot,Y_1),\ldots,k_\calY(\cdot,Y_n)\right)^\rmT$, and $R_{X|Y} = \Lambda G_Y((\Lambda G_Y)^2+\delta I_n)^{-1}\Lambda$, 
where $I_n$ is the identity matrix of size $n$ and $\epsilon,\delta\in\R$ are heuristically set regularization parameters.
Note that $\mathbbm{1}_A$ stands for the indicator function of a set $A$ described as
\begin{equation*}
  \mathbbm{1}_A(t) := \begin{cases}
    \ \ 1 & (t\in A) \\
    \ \ 0 & (t\notin A)\end{cases}.
\end{equation*}
Following Theorem \ref{thm:kbr-prop4}, the posterior kernel mean given a test value $y\in\calY$ is estimated by
\begin{equation*}
  \widehat{m}_{Q_{\calX|y}} = \bs{k}_X^\rmT R_{X|Y}\bs{k}_Y(y).
\end{equation*}
Here, we estimate the posterior probabilities for classes given a test value $y\in\calY$ by
\begin{equation*}
  \left(\begin{array}{c}
    \widehat{Q}_{\calX|y}(C_1)\\
    \vdots\\
    \widehat{Q}_{\calX|y}(C_g)
  \end{array}\right)
  = \left(\begin{array}{c}
    \left<\mathbbm{1}_{\{C_1\}},\,\widehat{m}_{Q_{\calX|y}}\right>_{\calH_\calX}\\
    \vdots\\
    \left<\mathbbm{1}_{\{C_g\}},\,\widehat{m}_{Q_{\calX|y}}\right>_{\calH_\calX}
  \end{array}\right)
  = DR_{X|Y}\bs{k}_Y(y).
\end{equation*}

\subsubsection{The algorithm of KBR2}
\noindent
Let $G_X^\dagger$ denote the Moore-Penrose generalized inverse matrix of $G_X$. Let us put $\widehat{\bs{\mu}}' = (\widehat{\mu}'_1, \ldots, \widehat{\mu}'_n)^\rmT = G_X^\dagger\widehat{\bs{m}}_\Pi$,
$\Lambda' = \diag(\,\widehat{\bs{\mu}}')$, and $R'_{X|Y} = \left(\Lambda' G_Y\right)^\dagger\Lambda'$.
Replacing $R_{X|Y}$ in \S\ref{subsubsec:4-1-2} for $R'_{X|Y}$, the posterior probabilities for classes given a test value $y\in\calY$ is estimated by
\begin{equation*}
  \left(\widehat{Q}_{\calX|y}(C_1),\ldots,\widehat{Q}_{\calX|y}(C_g)\right)^\rmT = DR'_{X|Y}\bs{k}_Y(y).
\end{equation*}

\subsection{Probabilistic predictions by the three classifiers}\label{subsec:4-2}
\begin{figure}[h!]
\centering
\includegraphics[width=0.6\textwidth]{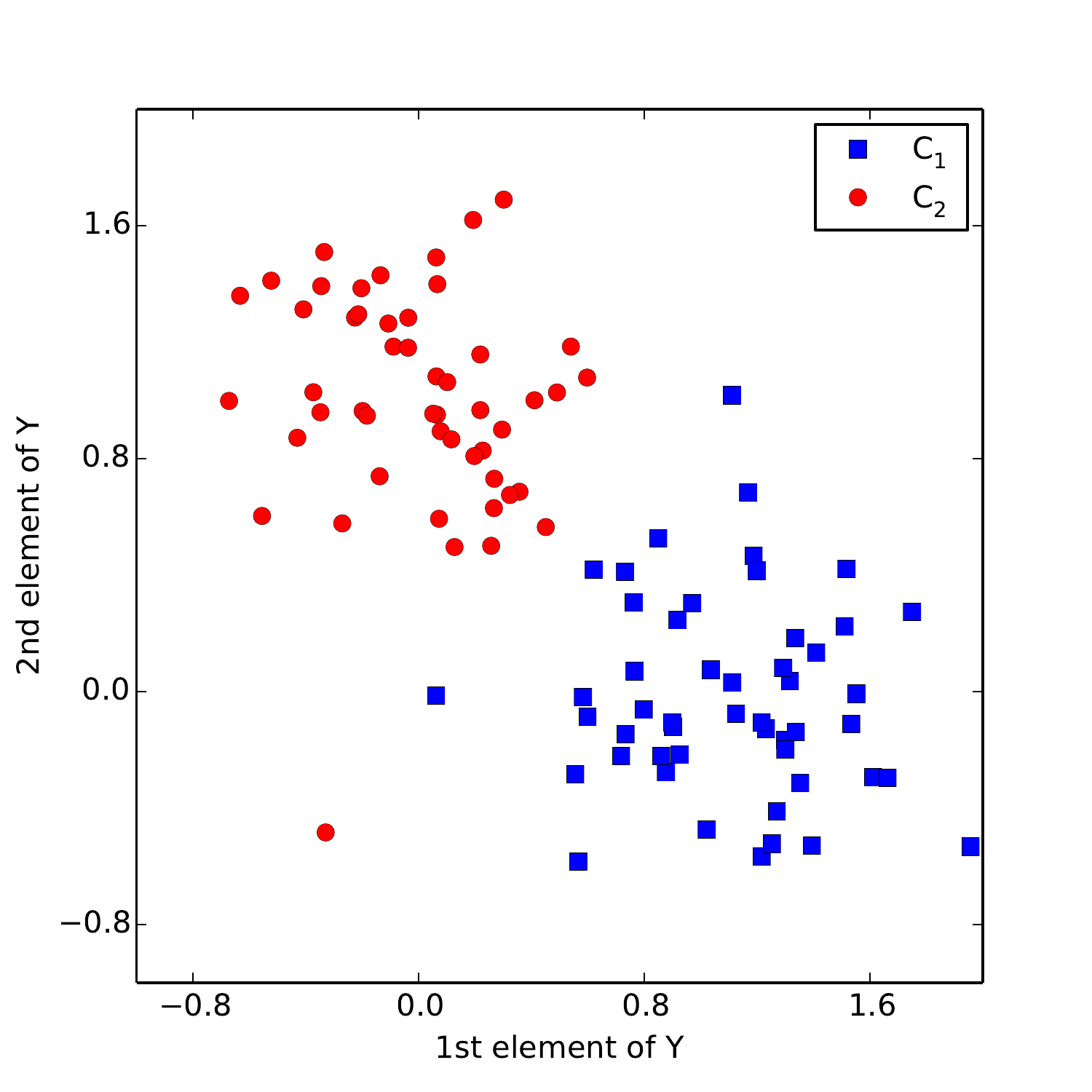}
\caption{Individual $Y$-values of a training sample}
\label{fig:1}
\end{figure}

\noindent
In this subsection, we apply the three classifiers defined in \S\ref{subsec:4-1} to a binary classification problem using computer-simulated data sets, where $\calX = \{ C_1, C_2 \}$ and $\calY = \R^2$.
In the first step, we independently generate 100 sets of training samples with each training sample being $\{(X_i, Y_i)\in\calX \times \calY\}_{i=1}^{100}$, where $X_{i} = C_1$ and $Y_i \sim \calN(M_1, S_1)$ if $1 \le i \le 50$, $X_{i} = C_2$ and $Y_i \sim \calN(M_2, S_2)$ if $51 \le i \le 100$, $M_1 = (1, 0)^\rmT$, $M_2 = (0, 1)^\rmT$, and $S_1 = S_2 = \diag(0.1, 0.1)$. 
Here, $\{Y_i\}_{i=1}^{50}$ and $\{Y_i\}_{i=51}^{100}$ are sampled i.i.d.\  from $\calN(M_1, S_1)$ and $\calN(M_2, S_2)$, respectively.
Individual $Y$-values of one of the training samples are plotted in Figure \ref{fig:1}.

\begin{figure}[h!]
\centering
\includegraphics[width=\textwidth]{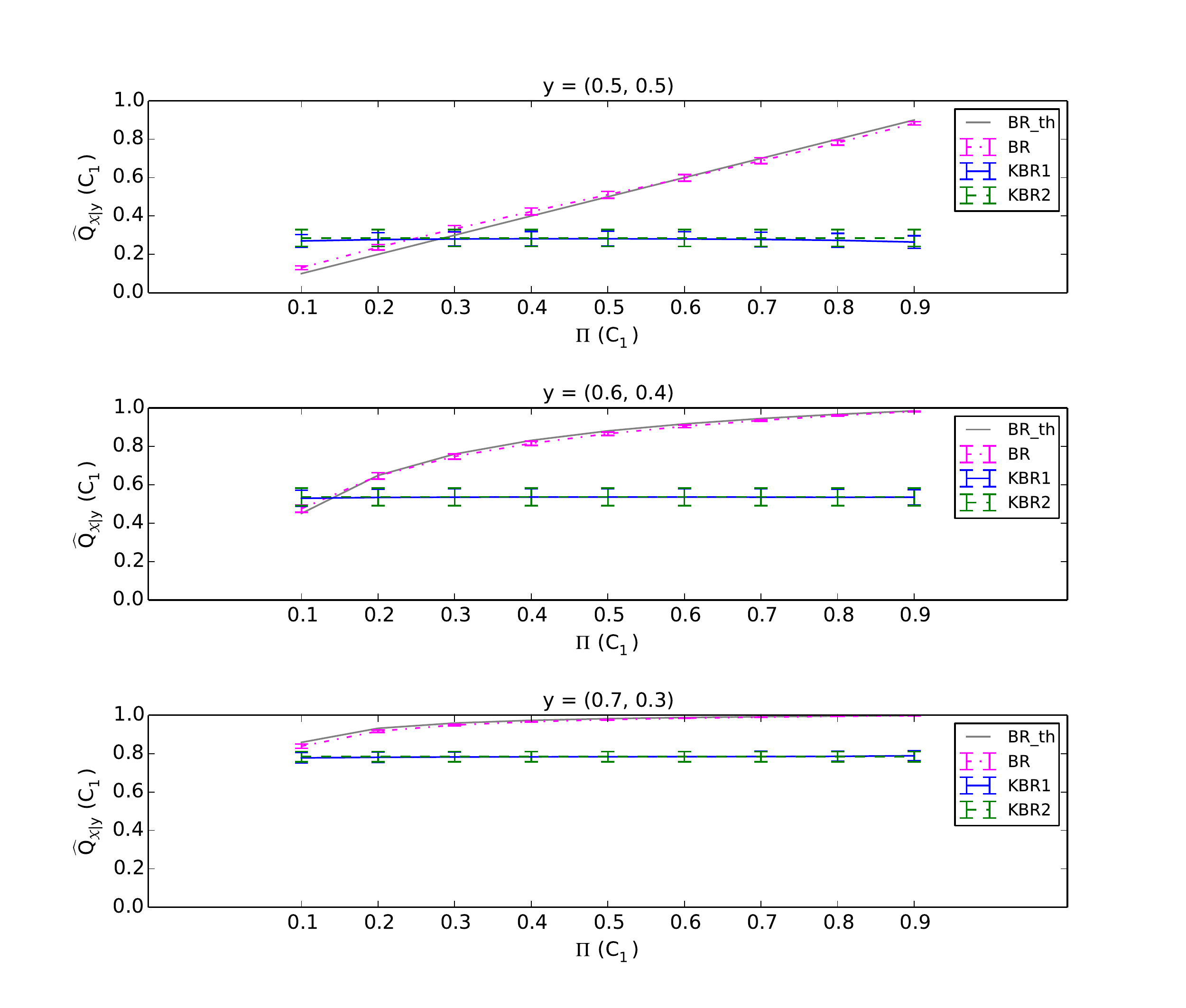}
\caption{The case $\epsilon=\delta=10^{-7}$}
\label{fig:2}
\end{figure}

\begin{figure}[h!]
\centering
\includegraphics[width=\textwidth]{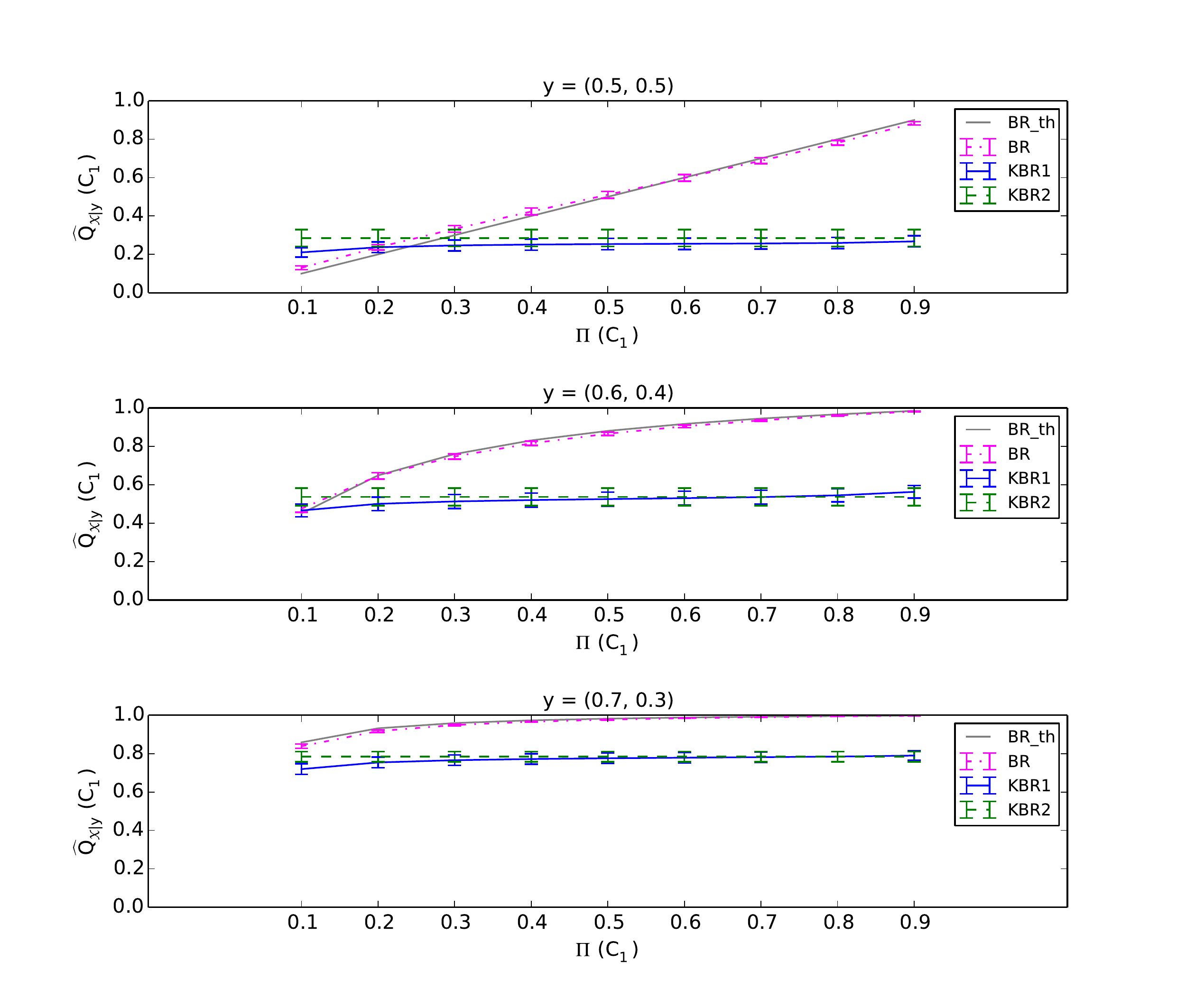}
\caption{The case $\epsilon=\delta=10^{-5}$}
\label{fig:3}
\end{figure}

\begin{figure}[h!]
\centering
\includegraphics[width=\textwidth]{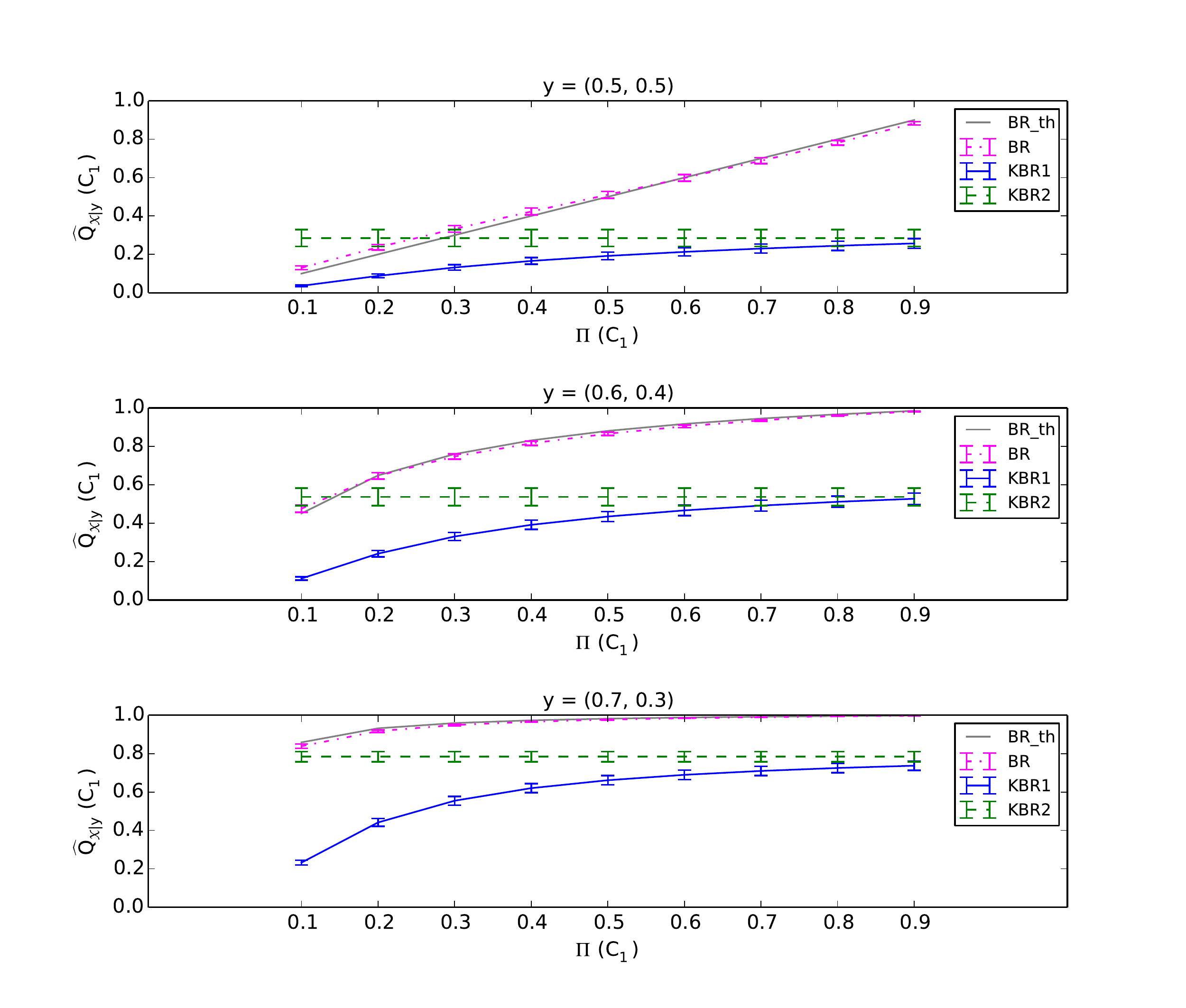}
\caption{The case $\epsilon=\delta=10^{-3}$}
\label{fig:4}
\end{figure}

\begin{figure}[h!]
\centering
\includegraphics[width=\textwidth]{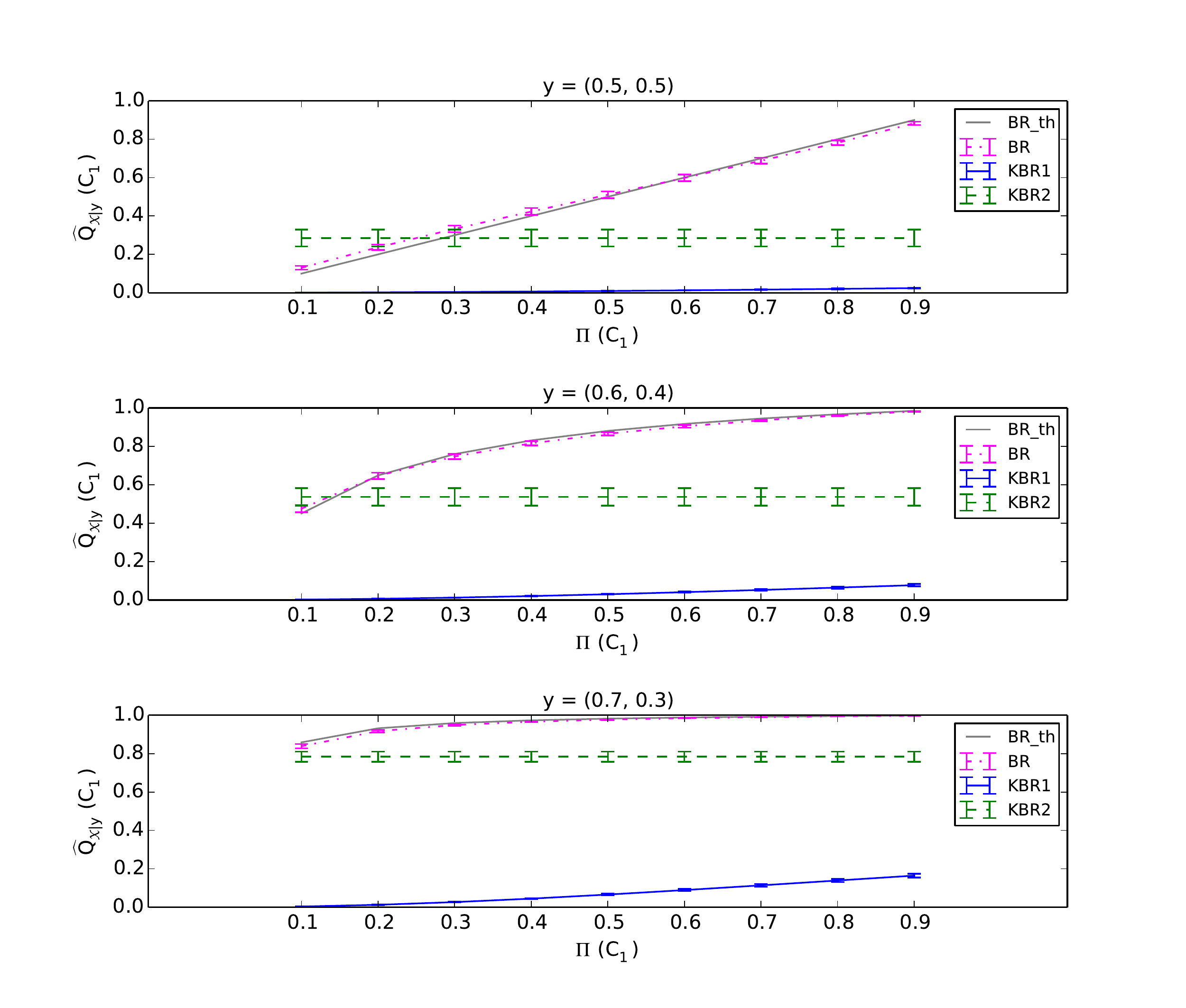}
\caption{The case $\epsilon=\delta=10^{-1}$}
\label{fig:5}
\end{figure}

With each of the 100 training samples and a simulated prior probability of $C_1$, or $\Pi(C_1)\in\{0.1, 0.2,\ldots,0.9\}$, the classifiers defined in \S\ref{subsec:4-1} estimate the posterior probability of $C_1$ given a test value $y\in\{(0.5, 0.5), (0.6, 0.4), (0.7, 0.3)\}$, that is, $Q_{\calX|y}(C_1)$.
Figures \ref{fig:2}-\ref{fig:5} show the mean (plus or minus standard error of the mean, SEM) of the 100 values of $\widehat{Q}_{\calX|y}(C_1)$ calculated by each of the classifiers, BR, KBR1, and KBR2.
Here we show the case where $\sigma$ in KBR1 and KBR2 is fixed to $0.1$, and the regularization parameters of KBR1 are set to be $\epsilon=\delta=10^{-7}$ (Figure \ref{fig:2}), $\epsilon=\delta=10^{-5}$ (Figure \ref{fig:3}), $\epsilon=\delta=10^{-3}$ (Figure \ref{fig:4}), and $\epsilon=\delta=10^{-1}$ (Figure \ref{fig:5}). In Figures \ref{fig:2}-\ref{fig:5}, BR\_th illustrates the theoretical result of BR, where $\widehat{M}_1$, $\widehat{M}_2$, $\widehat{S}_1$, and $\widehat{S}_2$ in BR are set to be $M_1$, $M_2$, $S_1$, and $S_2$, respectively.

Consistent to \S\ref{subsec:3-1}, $\widehat{Q}_{\calX|y}(C_1)$ calculated by KBR1 is poorly influenced by $\Pi(C_1)$ compared with that by BR when $\epsilon$ and $\delta$ are set to be small (see Figures \ref{fig:2} and \ref{fig:3}).
In addition, $\widehat{Q}_{\calX|y}(C_1)$ calculated by KBR2 also seems to be uninfluenced by $\Pi(C_1)$. 
When $\epsilon$ and $\delta$ are set to be larger, the effect of $\Pi(C_1)$ on  $\widehat{Q}_{\calX|y}(C_1)$ becomes apparent in KBR1, however, the value of $\widehat{Q}_{\calX|y}(C_1)$ becomes too small (see Figures \ref{fig:4} and \ref{fig:5}).
These results suggest that in kernel Bayes' rule, the posterior does not depend on the prior if $\epsilon$ and $\delta$ are negligible, which might be a contradiction to the nature of Bayes' theorem.
Moreover, even though the prior affects the posterior when $\epsilon$ and $\delta$ become larger, the posterior seems too much dependent on $\epsilon$ and $\delta$, which are initially defined just for the regularization of matrices.

We have also tested all possible combinations of the following values for the parameters in KBR1 and/or KBR2: $\epsilon\in\{10^{-1}, 10^{-3}, 10^{-5}, 10^{-7}, 10^{-9}, 10^{-11}, 10^{-13}, 10^{-15}\}$, $\delta\in\{10^{-1}, 10^{-3}, 10^{-5}, 10^{-7}, 10^{-9}, 10^{-11}, 10^{-13}, 10^{-15}\}$, and $\sigma\in\{0.01,0.1,1,10,100\}$.
All the experimental results have been evaluated in a similar manner as above, and none of the results are found to be reasonable in the context of Bayesian inference.

\section{Conclusions}\label{sec:5}
\noindent
One of the important features of Bayesian inference is that it provides a reasonable way of updating the probability for a hypothesis as additional evidence is acquired.
Kernel Bayes' rule has been expected to enable Bayesian inference in RKHS. 
In other words, the posterior kernel mean has been considered to be reasonably estimated by kernel Bayes' rule, given kernel mean expressions of the prior and likelihood.
What is ``reasonable" depends on circumstances, however, some of the results in this paper seem to show obviously unreasonable aspects of kernel Bayes' rule, at least in the context of Bayesian inference.

First, as shown in \S\ref{subsec:3-1}, when $\Lambda$ and $G_Y$ are non-singular matrices and so we set $\delta = 0$, the posterior kernel mean $\widehat{m}_{Q_{\calX|y}}$ is entirely unaffected by the prior distribution $\Pi$ on $\calX$.
This means that, in Bayesian inference with kernel Bayes' rule, prior beliefs are in some cases completely neglected in calculating the kernel mean of the posterior distribution.
Numerical evidence is also presented in \S\ref{subsec:4-2}.
When the regularization parameters $\epsilon$ and $\delta$ are set to be small, the posterior probability calculated by kernel Bayes' rule (KBR1) is almost unaffected by the prior probability in comparison with that by conventional Bayes' rule (BR).
Consistently, when the regularized inverse matrices in KBR1 are replaced for the Moore-Penrose generalized inverse matrices (KBR2), the posterior probability is also uninfluenced by the prior probability, which seems to be unsuitable in the context of Bayesian updating of a probability distribution.

Second, as discussed in \S\ref{subsec:3-2} and \S\ref{subsec:4-2}, the posterior estimated by kernel Bayes' rule considerably depends upon the regularization parameters $\epsilon$ and $\delta$, which are originally introduced just for the regularization of matrices.
A cross-validation approach is proposed in \cite{fukumizu13} to search for the optimal values of the parameters.
However, theoretical foundations seem to be insufficient for the correct tuning of the parameters.
Furthermore, in our experimental settings, we are not able to obtain a reasonable result  using any combination of the parameter values, suggesting the possibility that there are no appropriate values for the parameters in general.
Thus, we consider it difficult to solve the problem that $C_{XX}$ and $C_{WW}$ are not surjective by just adding regularization parameters.

Third, as shown in \S\ref{subsec:3-3}, the assumption that conditional expectation functions are included in the RKHS does not hold in general.
Since this assumption is necessary for most of the theorems in \cite{fukumizu13}, we believe that the assumption itself may need to be reconsidered.

In summary, even though current research efforts are focused on the application of kernel Bayes' rule \cite{fukumizu13, kanagawa14}, it might be necessary to reexamine its basic framework of combining new evidence with prior beliefs.

\section{Appendix}\label{sec:6}
\noindent
In this section, we provide some proofs for \S\ref{sec:2} and \S\ref{sec:3}.
\subsection{Estimation of $C_{ZW}$ and $C_{WW}$}\label{sec:rev_prf}
\noindent
Here we give the proof of Proposition \ref{prop:rev}.
\begin{prf}
Let $\widehat{C}_{XX}$, $\widehat{C}_{(YX)X}$, and $\widehat{C}_{(YY)X}$ denote the estimates of $C_{XX}$, $C_{(YX)X}$, and $C_{(YY)X}$, respectively.
We define the estimates of $m_{(ZW)}$ and $m_{(WW)}$ as
\begin{equation*}
  \widehat{m}_{(ZW)} = \widehat{C}_{(YX)X}\widehat{C}_{XX}^{-1}\widehat{m}_\Pi
  \quad\mbox{and}\quad
  \widehat{m}_{(WW)} = \widehat{C}_{(YY)X}\widehat{C}_{XX}^{-1}\widehat{m}_\Pi, 
\end{equation*}
and put $h = \widehat{C}_{XX}^{-1}\widehat{m}_\Pi\in\calH_\calX$.
According to Equations (\ref{eq:2-5}), for any $f\in\calH_\calX$ and $g\in\calH_\calY$,
\begin{align*}
  &\left\langle \widehat{m}_{(ZW)}, g\otimes f\right\rangle_{\calH_\calY\otimes\calH_\calX}
  = \left\langle \widehat{C}_{(YX)X}h, g\otimes f\right\rangle_{\calH_\calY\otimes\calH_\calX}
  = \widehat{E}\left[f(X)g(Y)h(X)\right] \\
  &= \frac{1}{n}\sum_{i=1}^nf(X_i)g(Y_i)h(X_i)
  = \left\langle \frac{1}{n}\sum_{i=1}^nh(X_i)k_\calX(\cdot, X_i)\otimes k_\calY(\cdot, Y_i),\, f\otimes g\right\rangle_{\calH_\calX\otimes\calH_\calY},
\end{align*}
where $\widehat{E}[\cdot]$ represents the empirical expectation operator.
Thus, from Remark \ref{prop:2-1}, 
\begin{equation}\label{eq:2-7}
  \widehat{C}_{ZW} = \widehat{m}_{(ZW)} = \frac{1}{n}\sum_{i=1}^nh(X_i)k_\calX(\cdot, X_i)\otimes k_\calY(\cdot, Y_i).
\end{equation}
Similarly, for any $g_1, g_2\in\calH_\calY$,
\begin{align*}
  &\left\langle \widehat{m}_{(WW)}, g_1\otimes g_2\right\rangle_{\calH_\calY\otimes\calH_\calY}
  = \left\langle \widehat{C}_{(YY)X}h, g_1\otimes g_2\right\rangle_{\calH_\calY\otimes\calH_\calY}
  = \widehat{E}\left[g_1(Y)g_2(Y)h(X)\right] \\
  &= \frac{1}{n}\sum_{i=1}^ng_1(Y_i)g_2(Y_i)h(X_i)
  = \left\langle \frac{1}{n}\sum_{i=1}^nh(X_i)k_\calY(\cdot, Y_i)\otimes k_\calY(\cdot, Y_i),\, g_1\otimes g_2\right\rangle_{\calH_\calY\otimes\calH_\calY}.
\end{align*}
Thus, from Remark \ref{prop:2-1}, 
\begin{equation}\label{eq:2-8}
  \widehat{C}_{WW} = \widehat{m}_{(WW)} = \frac{1}{n}\sum_{i=1}^nh(X_i)k_\calY(\cdot, Y_i)\otimes k_\calY(\cdot, Y_i).
\end{equation}
Next, we will derive $h(X_1), \ldots, h(X_n)$.
Since $C_{XX}$ is a self-adjoint operator,
\begin{align*}
  \left\langle h, \widehat{C}_{XX}f\right\rangle_{\calH_\calX}
  &= \left\langle \widehat{C}_{XX}h, f\right\rangle_{\calH_\calX}
  =\left\langle \widehat{m}_\Pi, f\right\rangle_{\calH_\calX} 
  = \sum_{j=1}^l\gamma_jf(U_j)
\end{align*}
for any $f\in\calH_\calX$. On the other hand, from Equations (\ref{eq:2-1}), 
\begin{equation*}
  \left\langle h, \widehat{C}_{XX}f\right\rangle_{\calH_\calX}
  = \widehat{E}\left[f(X)h(X)\right]
  = \frac{1}{n}\sum_{i=1}^nf(X_i)h(X_i)
\end{equation*}
for any $f\in\calH_\calX$.
Hence, we have
\begin{equation}\label{eq:2-9}
  \sum_{j=1}^l\gamma_jf(U_j) = \frac{1}{n}\sum_{i=1}^nf(X_i)h(X_i)
\end{equation}
for any $f\in\calH_\calX$.
Replacing $f$ in Equation (\ref{eq:2-9}) for $k_\calX(X_1, \cdot),\ldots,k_\calX(X_n, \cdot)\in\calH_\calX$, we have
\begin{equation}\label{eq:2-10}
  \left(\begin{array}{ccc}
    k_\calX(X_1, U_1) & \cdots & k_\calX(X_1, U_l) \\
    \vdots & \ddots & \vdots \\
    k_\calX(X_n, U_1) & \cdots & k_\calX(X_n, U_l)
  \end{array}\right)
  \left(\begin{array}{c}
    \gamma_1 \\
    \vdots \\
    \gamma_l
  \end{array}\right)
  = \frac{1}{n}G_X \left(\begin{array}{c}
    h(X_1) \\
    \vdots \\
    h(X_n)
  \end{array}\right).
\end{equation}
Using Equation (\ref{eq:2-6}), the left hand side of Equation (\ref{eq:2-10}) is given by
\begin{equation*}
  \left(\begin{array}{c}
  \left\langle\sum_{j=1}^l\gamma_jk_\calX (\cdot, U_j),\, k_\calX(\cdot, X_1)\right\rangle_{\calH_\calX} \\
  \vdots \\
  \left\langle\sum_{j=1}^l\gamma_jk_\calX (\cdot, U_j),\, k_\calX(\cdot, X_n)\right\rangle_{\calH_\calX}
  \end{array}\right)
  = \left(\begin{array}{c}
  \sum_{j=1}^l\gamma_jk_\calX (X_1, U_j) \\
  \vdots \\
  \sum_{j=1}^l\gamma_jk_\calX (X_n, U_j)
  \end{array}\right)
  = \left(\begin{array}{c}
    \widehat{m}_\Pi(X_1) \\
    \vdots \\
    \widehat{m}_\Pi(X_n)
  \end{array}\right).
\end{equation*}
Therefore, we have
\begin{equation*}
  \frac{1}{n}\left(\begin{array}{c}
    h(X_1) \\
    \vdots \\
    h(X_n)\end{array}\right)
  = G_X^{-1}\left(\begin{array}{c}
    \widehat{m}_\Pi(X_1) \\
    \vdots \\
    \widehat{m}_\Pi(X_n)\end{array}\right)
  \approx \left(G_X+n\epsilon I\right)^{-1}\widehat{\bs{m}}_\Pi
  = \widehat{\bs{\mu}}.
\end{equation*}
Replacing $\frac{1}{n}(h(X_1),\ldots,h(X_n))^\rmT$ for $\widehat{\bs{\mu}} = (\widehat{\mu}_1,\ldots,\widehat{\mu}_n)^\rmT$, Equations (\ref{eq:2-7}) and (\ref{eq:2-8}) become
\begin{equation*}
  \widehat{C}_{ZW} = \sum_{i=1}^n\widehat{\mu}_ik_\calX(\cdot,X_i)\otimes k_\calY(\cdot,Y_i)
  \quad\mbox{and}\quad
  \widehat{C}_{WW} = \sum_{i=1}^n\widehat{\mu}_ik_\calY(\cdot,Y_i)\otimes k_\calY(\cdot,Y_i),
\end{equation*}
respectively.\qed
\end{prf}

\subsection{Non-singularity of $G_Y$ and $\Lambda$}\label{sec:nonsing}
\noindent
Here we show that the assumption in \S\ref{subsec:3-1} holds under reasonable conditions.

\begin{defn}
Let $f$ be a real-valued function defined on a non-empty open domain $\dom(f)\subseteq\R^d$. 
We say that $f$ is analytic if $f$ can be described by a Taylor expansion on 
a neighborhood of each point of $\dom(f)$. 
\end{defn}

\begin{prop}\label{prop:non-sing} 
Let $k$ be a positive definite kernel on $\R^d$. 
Let $\nu$ be a probability measure on $\R^d$ which is absolutely continuous with respect to Lebesgue measure. 
Assume that $k$ is an analytic function on $\R^d \times \R^d$ and that the RKHS corresponding to $k$ is infinite dimensional. 
Then for any i.i.d.\  random variables $X_1,  X_2, \ldots, X_n$ with the same 
distribution $\nu$, the Gram matrix $G_X = (k(X_{i}, X_{j}))_{1\le i, j \le n}$ is non-singular almost surely with respect to $\nu^n = \nu\times\nu\times\cdots\times\nu\ \mbox{(n times)}$.  
\end{prop}

\begin{prf}
Let us put $f(x_1, x_2, \ldots, x_n) := \det (k(x_{i}, x_{j}))_{1\le i, j \le n}$. 
Since the RKHS corresponding to $k$ is infinite dimensional, there are $\xi_1, \xi_2, \ldots, \xi_n \in \R^d$ such that $\{ k(\cdot, \xi_i) \}_{1\le i \le n}$ are linearly independent. 
Then $f(\xi_1, \xi_2, \ldots, \xi_n) \neq 0$ and hence 
$f$ is a non-zero analytic function. 
Note that any non-trivial 
subvarieties of the euclidean spaces defined by analytic functions 
have Lebesgue measure zero. By this fact, the subvariety
\begin{align*}
  \mathcal{V}(f) := \left\{\begin{array}{c|c}(x_1, x_2, \ldots, x_n) \in (\R^d)^n & f (x_1, x_2, \ldots, x_n) = 0\end{array}\right\}\subset (\R^d)^n
\end{align*}  
has Lebesgue 
measure zero. Since $\nu$ is absolutely continuous, $\nu^n(\mathcal{V}(f)) = 0$. This completes the proof. 
\qed 
\end{prf}

From Proposition \ref{prop:non-sing}, we easily obtain the following corollary. 

\begin{cor}
Let $k$ be a Gaussian kernel on $\R^d$ and let $X_1,  X_2, \ldots, X_n$ be i.i.d.\  random variables with the same normal distribution on $\R^d$. 
Then the Gram matrix $G_X = (k(X_{i}, X_{j}))_{1\le i, j \le n}$ is non-singular almost surely. 
\end{cor}

\begin{prop}\label{prop:munon-zero}
Let $k$ be a positive definite kernel on $\calX = \R^d$, $\nu$ a probability measure on $\calX$ which is absolutely continuous with respect to Lebesgue measure. 
Assume that $k$ is an analytic function on $\calX \times \calX$ and that the RKHS $\calH$ corresponding to $k$ is infinite dimensional. 
Then for any $(\epsilon, \gamma_1, \gamma_2, \ldots, \gamma_l, U_1, U_2, \ldots, U_l) \in \R_{+}\times \R^l \times (\R^d)^l$ except Lebesgue measure zero, 
and for any  i.i.d.\  random variables $X_1,  X_2, \ldots, X_n$ with the same 
distribution $\nu$, each $\mu_i$ for $i=1,2,\ldots,n$ is non-zero almost surely, where 
$(\mu_1, \mu_2, \ldots, \mu_{n})^{T} = (G_X + n \epsilon I_n)^{-1} \widehat{\bs{m}}_{\Pi}$, $\widehat{\bs{m}}_{\Pi} = ( \widehat{m}_{\Pi}(X_1), \widehat{m}_{\Pi}(X_2), \ldots, \widehat{m}_{\Pi}(X_n))^{T}$, and $\widehat{m}_{\Pi}(\cdot) = \sum_{j=1}^{l} \gamma_j k(\cdot, U_j)$.
Here $\R_+$ denotes the set of positive real numbers.
\end{prop}

\begin{prf}
Let us put ${\cal S} := \R_{+}\times \R^l \times (\R^d)^l$, ${\cal T} := \calX^n \times {\cal S}$, and
\begin{equation*}
  f_i(x_1, x_2, \ldots, x_{n}, \epsilon, \gamma_1, \gamma_2, \ldots, \gamma_l, U_1, U_2, \ldots, U_l) := \mu_i \quad (i=1,2,\ldots, n)
\end{equation*}
for $(x_1, x_2, \ldots, x_{n}) \in \calX^n$ and $(\epsilon, \gamma_1, \gamma_2, \ldots, \gamma_l, U_1, U_2, \ldots, U_l) \in {\cal S}$.
We can verify that $G_X + n\epsilon I_n = (k(x_{i}, x_{j}))_{1\le i, j \le n}+n\epsilon I_n$ is non-singular almost everywhere on ${\cal T}$ in the same way as in the proof of Proposition \ref{prop:non-sing}. 
Let us define a closed measure-zero set $\mathcal{V} := \{(x_1, x_2, \ldots, x_{n}, \epsilon, \gamma_1, \gamma_2, \ldots, \gamma_l, U_1, U_2, \ldots, U_l)\in\mathcal{T} \mid \det(G_X + n\epsilon I_n) = 0  \}\subset {\cal T}$. 
Then $f_i$ is defined on $\mathcal{T}\setminus \mathcal{V}$ for each $i\in\{1,2,\ldots,n\}$. 
Using Cramer's rule, 
\begin{equation*}
  \mu_i = \frac{\det\left(\eta_1,\eta_2,\ldots, \eta_{i-1},\widehat{\bs{m}}_\Pi,\eta_{i+1},\ldots,\eta_n\right)}{\det\left(G_X+n\epsilon I_n\right)},
\end{equation*}
where $\eta_m$ stands for the $m$-th column vector of $G_X + n\epsilon I_n$. 
Here we denote by $g_i$ the numerator of $\mu_i$, that is, $g_i = \mu_i 
\det( G_X + n\epsilon I_n )$. 
Let us choose $\xi_1, \xi_2, \ldots, \xi_n \in \calX$ such that $\{ k(\cdot, \xi_i) \}_{1 \le i \le n}$ are linearly independent in $\calH$.
It is easy to see that $g_i(\xi_1, \xi_2, \ldots, \xi_n, \ast)$ is a non-zero 
analytic function of $\ast$ on ${\cal S}$.
Indeed, if $\epsilon\!\rightarrow\!+0$, $U_1 = \xi_i$, $\gamma_1 = 1$, and $\gamma_2=\gamma_3=\cdots=\gamma_l = 0$, then $g_i \rightarrow \det(\langle k(\cdot,\xi_i), k(\cdot,\xi_j) \rangle_\calH)_{1\le i, j\le n} \ne 0$.
Hence $\mathcal{Z}_i := \{  \ast \in {\cal S} \mid g_i(\xi_1, \xi_2, \ldots, \xi_n, \ast) = 0 \}$ is a closed subset of ${\cal S}$ with Lebesgue measure zero for each $i\in\{1,2,\ldots,n\}$. 
Thus, since $g_i(\ast, \epsilon, \gamma_1, \gamma_2, \ldots, \gamma_l, U_1, U_2, \ldots, U_l)$ is a non-zero analytic function of $\ast$ on $\calX^n$ for any  
$(\epsilon, \gamma_1, \gamma_2, \ldots, \gamma_l, U_1, U_2, \ldots, U_l) \in {\cal S}\setminus (\cup_{i=1}^{n} \mathcal{Z}_i)$, 
\begin{align*}
  \mathcal{F}_i := \left\{\begin{array}{c|c}\ast \in \calX^n & 
  g_i\left(\ast, \epsilon, \gamma_1, \gamma_2, \ldots, \gamma_l, U_1, U_2, \ldots, U_l\right)
  = 0\end{array}\right\}  
\end{align*} 
is a closed subset of $\calX^n$ with Lebesgue measure zero for each $i\in\{1,2,\ldots,n\}$. 
Therefore $\mu_i = f_i(\ast, \epsilon, \gamma_1, \gamma_2, \ldots, \gamma_l, U_1, U_2, \ldots, U_l)$ is non-zero almost surely on $\calX^n$ for any $(\epsilon, \gamma_1, \gamma_2, \ldots, \gamma_l, U_1, U_2, \ldots, U_l) \in {\cal S}\setminus (\cup_{i=1}^{n} \mathcal{Z}_i)$ because the subset $\{ \ast \in \calX^n \mid f_i(\ast, \epsilon, \gamma_1, \gamma_2, \ldots, \gamma_l, U_1, U_2, \ldots, U_l)=0 \}$ 
is contained in $\mathcal{F}_i$ for each $i\in\{1,2,\ldots,n\}$. 
This completes the proof. \qed
\end{prf}

The following corollary directly follows from Proposition \ref{prop:munon-zero}. 

\begin{cor}
Let $k$ be a Gaussian kernel on $\R^d$ and let $X_1,  X_2, \ldots, X_n$ be i.i.d.\  random variables with the same normal distribution on $\R^d$. All other notations are as in Proposition \ref{prop:munon-zero}. Then $\Lambda := \diag(\mu_1, \mu_2, \ldots, \mu_n)$ is non-singular almost surely for any $(\epsilon, \gamma_1, \gamma_2, \ldots, \gamma_l, U_1, U_2, \ldots, U_l)\in \R_{+}\times \R^l \times (\R^d)^l$ except for those in a set of Lebesgue measure zero.
\end{cor}

\subsection{Non-surjectivity of $C_{XX}$ and $C_{WW}$}\label{sec:nonsur}
\noindent
The covariance operators $C_{XX}$ and $C_{WW}$ are not surjective in general. 
This can be verified by the fact that they are compact operators.
(If the operators are surjective on the corresponding RKHS which is infinite-dimensional, then they cannot be compact because of the open mapping theorem.)
Here we present some easy examples where $C_{XX}$ and $C_{WW}$ are not surjective.
Let us consider for simplicity the case $\calX=\R$. Let $X$ be a random variable on $\R$ with a normal distribution $\calN(\mu, \sigma_0^2)$. 
We prove that $C_{XX}$ is not surjective under the usual assumption that the positive definite kernel on $\R$ is Gaussian.
In order to demonstrate this, we use the symbols defined in \S\ref{subsec:3-3} and several proven results on function spaces and Fourier transforms (see \cite{rudin87}, for example). Note that the following three propositions are introduced without proofs.

\begin{prop}\label{prop:FT2} 
Let us put $f(x) = \exp(-(ax^2+bx+c))$ for $a, b, c \in \R$, where $a > 0$. 
Then
\begin{equation*}
\hat{f}(t) = \frac{1}{\sqrt{2a}} \exp\left( -\frac{t^2 - 2\sqrt{-1}bt - b^2+ 4ac}{4a} \right).  
\end{equation*}
\end{prop} 

\begin{prop}\label{prop:FTconjugate} 
For $f\in L^2(\R, \C)$, $\hat{\overline{f}}(t) = \overline{\hat{f}(-t)}$ almost everywhere.
In particular, if $f\in L^2(\R)$, then $\overline{\hat{f}(t)} = \hat{f}(-t)$ almost everywhere.
\end{prop} 

\begin{prop}\label{prop:fa1}
For $f\in L^2(\R, \C)$, put $f_a(x) := f(x-a)$. 
Then $\hat{f}_a(t) = \exp\left(-\sqrt{-1}at\right)\hat{f}(t)$. 
\end{prop}

\begin{defn}
Let $p(\cdot)$ denote the density function of the normal distribution $\calN(\mu, \sigma_{0}^2)$ on $\R$, that is, 
\begin{equation*}
p(\cdot) = \frac{1}{\sqrt{2\pi} \sigma_{0}} \exp \left( -\frac{(\cdot-\mu)^2}{2\sigma_{0}^2} \right). 
\end{equation*}
Let $X$ be a random variable on $\R$ with $\calN(\mu, \sigma_{0}^2)$. 
The linear operator $C_{XX} : \calH_{G} \to \calH_{G}$ is 
defined by $\langle C_{XX}f, g\rangle_{\calH_G} = E[f(X)g(X)]$ for any $f, g \in \calH_{G}$, which is also described as    
\begin{equation*}
(C_{XX}f)(\cdot) = \int_{-\infty}^{\infty} f(x) k(\cdot, x) p(x) \: dx  
\end{equation*}
for any $f \in \calH_G $. 
\end{defn} 

\begin{prop}
  If $f, g\in\calH_{G}$, then $\left<f, g \right>_{\calH_G}\in \R$
\end{prop}

\begin{prf}
From Proposition \ref{prop:FTconjugate}, $\overline{\hat{f}(t)} = \hat{f}(-t)$ and   
$\overline{\hat{g}(t)} = \hat{g}(-t)$ for any $f, g\in\calH_G $. 
Then, using Equation (\ref{eq:IP1}), we have 
\begin{align*}
  \overline{\left<f, g \right>_{\calH_G}} 
  &= \overline{\inftyint  \hat{f}(t) \overline{\hat{g}(t)} \exp \left(\frac{\sigma^2}{2} t^2 \right)\: dt}
  = \inftyint  \overline{\hat{f}(t)} \hat{g}(t) \exp \left(\frac{\sigma^2}{2} t^2 \right)\: dt \\
  &= \inftyint  \hat{f}(-t) \hat{g}(t) \exp \left(\frac{\sigma^2}{2} t^2 \right)\: dt 
  = \inftyint  \hat{f}(t) \hat{g}(-t) \exp \left(\frac{\sigma^2}{2} t^2 \right)\: dt \\
  &= \inftyint  \hat{f}(t) \overline{\hat{g}(t)} \exp \left(\frac{\sigma^2}{2} t^2 \right)\: dt 
  = \left<f, g \right>_{\calH_G}.
\end{align*}
Therefore, $\left<f, g \right>_{\calH_G} \in \R$. \qed
\end{prf}

\begin{prop} \label{prop:Hrc1} 
If $f \in\calH_{G}(\R, \C)$, then  $\overline{f}\in \calH_{G}(\R,\C)$.
\end{prop}

\begin{prf}
From Proposition \ref{prop:FTconjugate}, $\hat{\overline{f}}(t) = \overline{\hat{f}(-t)}$ 
for $f \in L^2(\R, \C)$.  
Then, using Equation (\ref{eq:IP1}), we have 
\begin{align*}
  \left\|\,\overline{f}\,\right\|_{\calH_G(\R, \C)}^2
  & = \left\langle\overline{f}, \overline{f}\right\rangle_{\calH_G(\R, \C)}\\ 
  &= \int_{-\infty}^{\infty} \left|\hat{\overline{f}}(t)\right|^2 \exp \left(\frac{\sigma^2}{2} t^2 \right) \: dt
  = \int_{-\infty}^{\infty} \left|\overline{\hat{f}(-t)}\right|^2 \exp \left(\frac{\sigma^2}{2} t^2 \right) \: dt \\
  &= \int_{-\infty}^{\infty} \left|\hat{f}(t)\right|^2 \exp \left(\frac{\sigma^2}{2} t^2 \right) \: dt 
  = \left\|f \right\|_{\calH_G(\R, \C)}^2 < \infty.
\end{align*}
Therefore, $\overline{f}\in\calH_{G}(\R, \C)$. \qed
\end{prf}

Here, we denote by $\rm{Re}$, $\rm{Im}$, and $\cl$ the real part of a complex number, the imaginary part of a complex number, and the closure operator, respectively.

\begin{cor} \label{cor:ReIm1} 
If $f\in\calH_{G}(\R, \C )$, then $\mathrm{Re}(f), \mathrm{Im}(f)\in \calH_{G}$.
\end{cor}

\begin{prf}
If $f\in\calH_{G}(\R, \C)$, then $\overline{f}\in\calH_{G} (\R, \C)$ by Proposition \ref{prop:Hrc1}. Hence we see that 
\begin{equation*}
  \mathrm{Re}(f) = \frac{f+\overline{f}}{2} \in \calH_{G}, \quad \mathrm{Im}(f) = \frac{f-\overline{f}}{2\sqrt{-1}} \in \calH_{G}. 
\end{equation*}
This completes the proof. \qed
\end{prf}

\begin{rmk}\label{rmk:hilbertdirectsum} 
If $f\in\calH_{G}(\R, \C)$, then there uniquely exist $f_1, f_2 \in \calH_{G}$ such that $f = f_1+\sqrt{-1}f_2$ by Corollary \ref{cor:ReIm1}. This means that 
$\calH_{G}(\R, \C) = \calH_{G}\oplus\sqrt{-1}\calH_{G}$, where $\oplus$ denotes the direct sum.
\end{rmk}

\begin{prop}\label{prop:HL1} 
For any $f\in L^2(\R, \C)$ and for any $\epsilon>0$, there exists $g\in\calH_{G}(\R, \C)$ such that $\|f-g \|_{2}< \epsilon$.
In other words, $\calH_{G}(\R,\C)$ is dense in $L^2(\R, \C)$. 
\end{prop}

\begin{prf} 
Let $C_0(\R, \C)$ denote the space of continuous complex-valued  
functions with compact support on $\R$.  
Let us define $\hat{\calH}_G(\R, \C)$ by 
\begin{equation*}
  \hat{\calH}_G \left(\R, \C  \right) := \left\{\begin{array}{c|c}
    h \in L^2(\R, \C) & \displaystyle \int_{-\infty}^{\infty}
    \left|h(t)\right|^2 \exp \left(\frac{\sigma^2}{2} t^2 \right)\:dt < \infty 
  \end{array}\right\}.  
\end{equation*}
Note that $\hat{\calH}_G (\R, \C)$ coincides with the image of $\calH_{G}(\R, \C)$ by the Fourier transform.  
Then, $C_0(\R, \C)\subset\hat{\calH}_G (\R, \C) \subset L^2(\R, \C)$ and $\cl(C_0(\R, \C)) = L^2(\R, \C)$.
Hence $\cl(\hat{\calH}_G (\R, \C)) = L^2(\R, \C)$. 
In other words, for any $f\in L^2(\R, \C)$ and for any $\epsilon > 0$, there exists $\hat{g}\in\hat{\calH}_G (\R, \C)$ such that 
$\|\hat{f}-\hat{g}\|_2<\epsilon$ because $\hat{f} \in L^2(\R, \C)$, which implies that there exists $g\in\calH_{G}(\R, \C)$ such that 
$\|f-g\|_2<\epsilon$. 
This completes the proof. \qed
\end{prf}

The following corollary has also been shown in Theorem 4.63 in \cite{steinwart08}.

\begin{cor} \label{cor:HL1} 
$\cl(\calH_{G}) = L^2(\R)$. 
\end{cor}

\begin{prf}
From Proposition \ref{prop:HL1}, for any $f\in L^2(\R)\subset L^2(\R, \C)$ and for any    
$\epsilon>0$, there exists $g\in\calH_{G}(\R, \C)$ such that $\|f-g\|_2<\epsilon$.   
By Remark \ref{rmk:hilbertdirectsum}, there exist $g_1, g_2\in\calH_{G}$ 
such that $g = g_1+\sqrt{-1}g_2$. Thus,
\begin{align*}
  \epsilon^2 &> \left\|f-g\right\|_{2}^2= \inftyint \left|f-g\right|^2\:dx 
  = \inftyint \left|(f-g_1)-\sqrt{-1}g_2\right|^2\:dx \\
  &\ge \inftyint \left|f-g_1\right|^2\:dx = \left\|f-g_1\right\|_2^2.
\end{align*}
Therefore $\left\|f-g_1\right\|_2 < \epsilon$.
This completes the proof. \qed
\end{prf}

\begin{defn}
Let us define $r, r_n\in L ^2(\R)$ as 
\begin{equation*}
  r(t) := \frac{1}{\left|\,t\,\right|} \mathbbm{1}_{(1,\infty)}(|\,t\,|), \quad  r_n(t) := \frac{1}{\left|\,t\,\right|} \mathbbm{1}_{(1,n)}(|\,t\,|),
\end{equation*}
where $\mathbbm{1}_{(1,\infty)}$ and $\mathbbm{1}_{(1,n)}$ denote the indicator functions of the intervals $(1,\infty)$ and $(1,n)$, respectively.
We also put $h_n:=\check{r_n}$ and $h:=\check{r}$. 
Note that $\displaystyle\lim_{n\to\infty} r_n = r\in L^2(\R)$, because
\begin{align*}
  \lim_{n\to\infty} \left\|r_n-r\right\|_2^2
  = 2\lim_{n\to\infty}\int_n^\infty\frac{1}{x^2}\: dx = 0.  
\end{align*}
\end{defn}

\begin{prop}
  $h_n, h\in L^2(\R)$.
\end{prop}
\begin{prf}
It is obvious that $h_n, h \in L^2(\R, \C)$. Since $r_n \in L^1(\R) \cap L^2(\R)$, we see that  
\begin{align*}
  \overline{h_n(x)} &= \overline{\check{r_n}(x)}
  = \overline{\frac{1}{\sqrt{2\pi}}\inftyint r_n(t)\exp\left(\sqrt{-1}tx\right)\:dt} \\
  &= \frac{1}{\sqrt{2\pi}}\inftyint r_n(t)\exp\left(-\sqrt{-1}tx\right)\:dt 
  = \frac{1}{\sqrt{2\pi}}\inftyint r_n(-t')\exp\left(\sqrt{-1}t'x\right)\:dt' \\
  &= \frac{1}{\sqrt{2\pi}}\inftyint r_n(t')\exp\left(\sqrt{-1}t'x\right)\:dt'
  = \check{r_n}(x)
  = h_n(x), 
\end{align*}
where $t' = -t$. Hence $h_n\in L^2(\R)$. On the other hand, 
\begin{eqnarray*}
  h(x) &=& \underset{n\to\infty\;\;}{\liminmean}\frac{1}{\sqrt{2\pi}}\int_{-n}^{n} r(t)\exp\left(\sqrt{-1}tx\right)\:dt \\
  &=& \underset{n\to\infty\;\;}{\liminmean}\frac{1}{\sqrt{2\pi}}\inftyint r_n(t)\exp\left(\sqrt{-1}tx\right)\:dt \\
  &=& \underset{n\to\infty\;\;}{\liminmean} h_n(x).
\end{eqnarray*}
Therefore $h\in L^2(\R)$.\qed
\end{prf}

Let us define $k_a(\cdot) := \sqrt{2\pi}\sigma k_G(\cdot, a)=\exp\left(-\frac{(\cdot\,-\,a)^2}{2\sigma^2}\right) \in \calH_G$ for $a\in\R$. 
Now, we prove that $k_a \notin \ran(C_{XX})$ for any $a \in \R$.
This implies that $C_{XX}$ is not surjective.   

\begin{prop}
For any $a \in \R$, $k_a\in\calH_{G}\setminus\ran (C_{XX})$.
\end{prop}

\begin{prf}
Suppose that there exists $g\in\calH_{G}$ such that $C_{XX}g=k_a$.
Then, for any $f\in\calH_{G}$,
\begin{equation} \label{eq:fh2} 
  \left<k_a,f\right>_{\calH_G} = \left<C_{XX}g, f\right>_{\calH_G}.
\end{equation}
Let us put $k(\cdot) = \sqrt{2\pi}\sigma k_G(\cdot, 0)=\exp\left(-\frac{(\cdot\,-\,0)^2}
  {2\sigma^2}\right)$. 
From Proposition \ref{prop:FT2}, $\hat{k}(t) = \sigma\exp\left(-\frac{\sigma^2}{2}t^2\right)$.  
Then, using Equation (\ref{eq:IP1}) and Proposition \ref{prop:fa1}, the left hand side of Equation (\ref{eq:fh2}) equals
\begin{align*}
  \inftyint\hat{k}_a(t)\overline{\hat{f}(t)}\exp\left(\frac{\sigma^2}{2}t^2\right)\:dt
  &= \inftyint\exp\left(-\sqrt{-1}at\right)\hat{k}(t)\overline{\hat{f}(t)}\exp\left(\frac{\sigma^2}{2}t^2\right)\:dt \\
  &= \sigma\inftyint\exp\left(-\sqrt{-1}at\right)\overline{\hat{f}(t)}\:dt.
\end{align*}
The right hand side of Equation (\ref{eq:fh2}) is equal to
\begin{align*}
  E\left[g(X)f(X)\right] &= \inftyint  g(x)f(x)p(x)\:dx 
  = \left<gp,f \right>_{L^2(\R)}.
\end{align*}
Thus, Equation (\ref{eq:fh2}) is equivalent to the following equation: 
\begin{equation} \label{eq:gph1}
  \left<gp,f \right>_{L^2(\R)} = \sigma\inftyint\exp\left(-\sqrt{-1}at\right)\overline{\hat{f}(t)}\:dt.
\end{equation}
Let us define $h_{n, a}(x) := h_n(x-a)$ and $h_a(x) = h(x-a)$.
Then $h_{n, a}, h_a\in L^2(\R)$. 
It is easy to see that 
$\|h_{n, a} - h_a\|_2 = \|h_n - h\|_2 = \|r_n - r\|_2\rightarrow 0$ as $n \to \infty$.
Hence $\displaystyle \lim_{n \to \infty} h_{n, a} = h_a$ in $L^2(\R)$. 
Since $\hat{h}_{n, a}(t) = \exp(-\sqrt{-1}at)\hat{h}_n(t)$ by Proposition \ref{prop:fa1}, we have 
\begin{align*}
  \inftyint\left|\hat{h}_{n, a}(t) \right|^2\exp\left(\frac{\sigma^2}{2}t^2\right)\:dt
  &= \inftyint\left|\hat{h}_n(t) \right|^2\exp\left(\frac{\sigma^2}{2}t^2\right)\:dt
  = \inftyint\left|r_n(t) \right|^2\exp\left(\frac{\sigma^2}{2}t^2\right)\:dt \\
  &= 2\int_1^n\frac{1}{t^2}\exp\left(\frac{\sigma^2}{2}t^2\right)\:dt
  < \infty,
\end{align*}
which indicates that $h_{n, a}\in\calH_{G}$.
Substituting $h_{n, a}$ for $f$, Equation (\ref{eq:gph1}) becomes
\begin{equation} \label{eq:gph2}
  \left<gp,h_{n, a} \right>_{L^2(\R)} = \sigma\inftyint\exp\left(-\sqrt{-1}at\right)\overline{\hat{h}_{n, a}(t)}\:dt.
\end{equation}
If $n$ goes to infinity, the left hand side of Equation (\ref{eq:gph2}) becomes 
$\langle gp, h_a \rangle_{L^2(\R)} \in\R$.
On the other hand, the right hand side of Equation (\ref{eq:gph2}) becomes
\begin{align*}
  \sigma\inftyint\exp\left(-\sqrt{-1}at\right)\overline{\exp(-\sqrt{-1}at)\hat{h}_n(t)}\:dt
  &= \sigma\inftyint\overline{\hat{h}_n(t)}\:dt \\
  &= \sigma\inftyint\overline{r_n(t)}\:dt \\
  &= 2\sigma\int_1^n\frac{1}{t}\:dt\rightarrow\infty\quad(n\rightarrow\infty).
\end{align*}
This is a contradiction. Therefore, there exists no $g\in\calH_{G}$ such that $C_{XX}g = k_a$. This completes the proof. \qed
\end{prf}

\section{Disclosures}
\noindent
The second author was partially supported by JSPS KAKENHI Grant Numbers 23540044, 15K04814.
The authors declare no other conflicts of interest.


\end{document}